# Alignment-Free Cross-Sensor Fingerprint Matching based on the Co-Occurrence of Ridge Orientations and Gabor-HoG Descriptor

Helala AlShehri[1], Muhammad Hussain[1], Hatim AboAlSamh[1], Senior Member, IEEE, Qazi Emad-ul-Haq[1], and Aqil M. Azmi [1]

[1] College of Computer and Information Sciences, King Saud University, Riyadh 11543, Kingdom of Saudi Arabia

**ABSTRACT** The existing automatic fingerprint verification methods are designed to work under the assumption that the same sensor is installed for enrollment and authentication (regular matching). There is a remarkable decrease in efficiency when one type of contact-based sensor is employed for enrolment and another type of contact-based sensor is used for authentication (cross-matching or fingerprint sensor interoperability problem,). The ridge orientation patterns in a fingerprint are invariant to sensor type. Based on this observation, we propose a robust fingerprint descriptor called the co-occurrence of ridge orientations (Co-Ror), which encodes the spatial distribution of ridge orientations. Employing this descriptor, we introduce an efficient automatic fingerprint verification method for cross-matching problem. Further, to enhance the robustness of the method, we incorporate scale based ridge orientation information through Gabor-HoG descriptor. The two descriptors are fused with canonical correlation analysis (CCA), and the matching score between two fingerprints is calculated using city-block distance. The proposed method is alignment-free and can handle the matching process without the need for a registration step. The intensive experiments on two benchmark databases (FingerPass and MOLF) show the effectiveness of the method and reveal its significant enhancement over the state-of-the-art methods such as VeriFinger (a commercial SDK), minutia cylinder-code (MCC), MCC with scale, and the thin-plate spline (TPS) model. The proposed research will help security agencies, service providers and law-enforcement departments to overcome the interoperability problem of contact sensors of different technology and interaction types.

## I. INTRODUCTION

Fingerprint matching is an active biometric research area and it is widely used for identify authentication. The existing methods for fingerprint matching are considered to be effective when the same sensor is employed for verification and enrollment. The advancement in fingerprint sensor technology and the growing number of fingerprint applications, matching the fingerprints of an individual captured with a variety of sensors has become a critical issue. Security agencies, service providers, and law-enforcement departments have vast fingerprint databases captured with a particular sensor. However, another sensor might be employed during verification and authentication. This introduced the fingerprint sensor interoperability problem. Fingerprint sensors [1] are based on various technologies like solid-state and ultrasound, which incorporate their own type of degradations in fingerprints, which makes the interoperability problem even more exigent. Cross-sensor fingerprint matching or sensor interoperability problem can be classified into two categories: (i) cross-matching between contact-based sensors of different technology and interaction types [7, 24] and (ii) cross-matching between contactless and contact sensors [26, 27]. In this paper, we address the first type of cross-sensor fingerprint matching problem.

Recent findings have emphasized the need to conduct research on cross matching methods [2], [3], [5]. However, this problem received the attention of few researchers. The main focus has been on the fusion of existing fingerprint-recognition methods [3], [24], the scaling of fingerprints [6]–[8], and nonlinear distortions [4], [25]. Despite these efforts, the contributions to crack the problem are marginal, and interoperability is still a challenge.

The fingerprints of a finger acquired with sensors of different types have identical ridge patterns, which play key role in discriminating fingerprints and serve as strong visual cues for identification. This observation motivated us to explore multiscale local ridge patterns and ridge orientation patterns for cross-matching of fingerprints. Taking it into consideration, we introduce a robust fingerprint descriptor that encodes distribution of ridge orientation patterns, and eventually an efficient fully automatic fingerprint authentication method. We argue that the descriptor can greatly reduce the effect of the sensor interoperability because it is based on ridge orientations, which are invariant to rotation and translation [9], [10] and the technology type of a sensor.

Furthermore, to incorporate scale based local ridge orientation information, we employ Gabor-HoG descriptor that enhances the effectiveness of the proposed method. The adopted descriptors emphasize on different fingerprint characteristics and to extract the most discriminative content, they are fused with canonical correlation analysis (CCA). The matching score of two fingerprints is calculated using city-block distance. The proposed method performs fingerprint matching without the need of alignment of minutia, which is an essential step in various methods in the fingerprint-matching literature. This registration free process reduces drastically the execution time of the matching step and makes the method simple and effective.

We evaluated the proposed method exhaustively on benchmark databases; the results indicate that it results in a better performance than the state-of-the-art methods such as VeriFinger (a commercial SDK), minutia cylinder-code (MCC), the thin-plate spline (TPS) model and MCC with scale.

Specifically, this work has the following major contributions.
1) An automatic fingerprint authentication method, which is effective for sensor interoperability problem. The method is an alignment-free approach, which reduces significantly the execution time of matching two fingerprints.
2) A novel fingerprint descriptor Co-Ror that represents a fingerprint as a spatial distribution of ridge orientations.
3) The fusion of Co-Ror and Gabor-HoG using CCA that turns out a robust fingerprint descriptor for cross matching problem.

The rest of the article is organized as follows. Section II gives an overview of the related work, and Section III presents high-level description of the proposed method and the detail of the feature extraction process. The evaluation protocol, empirical results are presented and discussed in Section IV. The conclusion has been drawn and the future work has been highlighted in Section V.

## II. RELATED WORK

Recent research demonstrated the significance of studying the effect of using different fingerprint sensors on automatic fingerprint-matching [5]. Jain and Ross [1] proved that the performance of a matching system decreases drastically when fingerprints are captured with two different sensors. Subsequently, Ross et al. [4] proposed a nonlinear calibration method that models the deformation of fingerprints employing the TPS model for registering a pair of fingerprints captured with different sensors.

Lugini et al. [2] addressed the sensor interoperability problem from a statistical point of view. They measured the degree of change in match scores when different sensors are employed for enrollment and verification. The study was performed using a large database captured from 494 participants with four different sensors, as well as the scanned ink-based fingerprints. The study's outcomes show that false non-match rates for fingerprint-matching systems are affected by the diversity of the capture devices but that false match rates are not. Mason et al. [11] proposed an approach to minimize the effects of low interoperability between optical sensors by combining some extracted fingerprint features with match scores using a classifier. The selected feature vector extracted from a fingerprint contained the following measures: average gray level, contrast, minutia count, quality measures, photo response non-uniformity (PRNU) noise, first-order statistics, mean of the orientation coherence matrix, and device ID. In addition, characteristics extracted from pairs of fingerprints, including alignment parameters and match scores, were used, and a tree-based scheme was implemented for classification.

Some researchers have studied the effect of adding a scale step to address cross-sensor matching. Ren et al. [8] introduced a scaling scheme, which is based on the average inter-ridge distance of a fingerprint and is used to compute the scale required to zoom-in onto two fingerprints to be compared. Zang et al. [6] developed a method to estimate the optimal scale between two fingerprints. First, the global coarse scale is calculated from the ridge distance map; then, the scale is computed using the histogram of the local refined scale between all corresponding minutia pairs. In [7], the state-of-the-art MCC matching system is modified with the addition of a scale step. In [24], the authors proposed a method for cross-sensor matching of fingerprints fusing three existing features. Though this method yields better performance than the state-of-the-art methods, it is time intensive because it involves alignment of minutia points.

### III. PROPOSED METHOD - CROSSVFINGER

An overview of the proposed method for cross-sensor fingerprint verification (CrossVFinger) is shown in Fig. 1. During the enrollment phase, fingerprints captured with sensor A are first preprocessed to reduce noise and to enhance their contrasts using the method proposed by Hong et al. [12]. Then, two types of descriptors (co-occurrence of ridge orientations (Co-Ror) and Gabor-HoG) are extracted and fused using CCA. Subsequently, the templates are stored in a template database.

During the authentication phase, a fingerprint captured with sensor B is first preprocessed; then, the descriptors are extracted and fused using CCA. The matching process is performed by computing the similarity between the fused descriptor and the respective template retrieved from the template database.

In the following sections, first we present the detail of the proposed fingerprint descriptor Co-Ror, which is the main contribution of this paper.

#### A. FINGERPRINT DESCRIPTION

In this section, we describe the feature extraction methods employed for fingerprint representation.

##### 1) CO-OCCURRENCE OF RIDGE ORIENTATIONS (CO-ROR)

Ridge orientation field of a fingerprint is not effected by translation, rotation and sensor type, and can be estimated with reasonable accuracy even from noisy fingerprints. It offers to develop a discriminative representation of a fingerprint, which is robust to changing sensors. Moreover, the distribution of ridge orientations contains important information that can be used to characterize the global shape of fingerprint ridge patterns. Therefore, we propose a fingerprint feature descriptor called the co-occurrence of ridge orientations (Co-Ror) to reveal certain information about the spatial distribution of the ridge orientation field of a fingerprint. The Co-Ror captures the spatial relationship between pairs of orientations by counting the frequency of co-occurrences of orientations.

The first step of extracting the Co-Ror is to compute the ridge orientation field of a fingerprint. Each element in the orientation field encodes the local orientation of fingerprint ridges [9]. Ridge orientation field is computed using the technique introduced in [12], where the dominant ridge orientation field is calculated by combining the gradient estimates within a window of size w × w centered at location (i, j):

$$\theta(i,j) = \frac{1}{2} \tan^{-1}\left(\frac{G_{yy}(i,j)}{G_{xx}(i,j)}\right) \quad (1)$$

where

$$G_{yy}(i,j) = \sum_{u=i-w/2}^{i+w/2} \sum_{v=j-w/2}^{j+w/2} 2(G_x(u,v)G_y(u,v))$$

$$G_{xx}(i,j) = \sum_{u=i-w/2}^{i+w/2} \sum_{v=j-w/2}^{j+w/2} (G^2_x(u,v) - G^2_y(u,v))$$

where $G_x$ and $G_y$ are the gradient magnitudes in the $x$ and $y$ directions, respectively, $\theta$ is in the range $[0, \pi]$. We applied Sobel operator to compute $G_x$ and $G_y$ because it has been extensively used to compute the gradients of fingerprints [13]–[15]. The Gaussian filter is applied to smooth the orientation of a window and to suppress noise as follows:

$$\theta'(i,j) = \frac{1}{2} \tan^{-1}\left(\frac{G(x,y)\sin(2\theta(i,j))}{G(x,y)\cos(2\theta(i,j))}\right) \quad (2)$$

where $G(x,y)$ is a Gaussian kernel.

Next, the orientation field is rotated so that the dominant orientation of the fingerprint is aligned with horizontal direction. To compute the dominant orientation, the histogram of orientations is created from the orientation fields and the peak of the histogram yields the dominant orientation ($\Theta$) of the fingerprint. Then, the ridge orientation field is rotated relative to the dominant orientation as follows:

$$\psi = \begin{cases} \theta - \Theta & \text{if } \theta \geq \Theta \\ \pi - \Theta + \theta & \text{otherwise.} \end{cases} \quad (3)$$

The $\psi$ represents the rotated ridge orientation field. A two-dimensional histogram $h_{d,\phi}$ is computed from $\psi$, where bin $h_{d,\phi}(i,j)$ represents the frequency of the co-occurrence of orientations $i$ and $j$ separated by distance $d$ in a direction specified by angle ϕ. We call this histogram a *Co-Ror matrix*, it involves two parameters: offset $d$ - the distance between orientations i & j and the direction ɸ of co-occurrence. As four directions are enough to determine the spatial structure of ridge orientation field, so for our analysis, we used four directions, ɸ = 0°, 45°, 90° and 135°, as shown in Fig. 2, resulting in four co-occurrence matrices. This choice has been validated in Section IV-B(1). The offset values are adopted based on inter-ridge distance; discussion is given in Section IV-B(1).

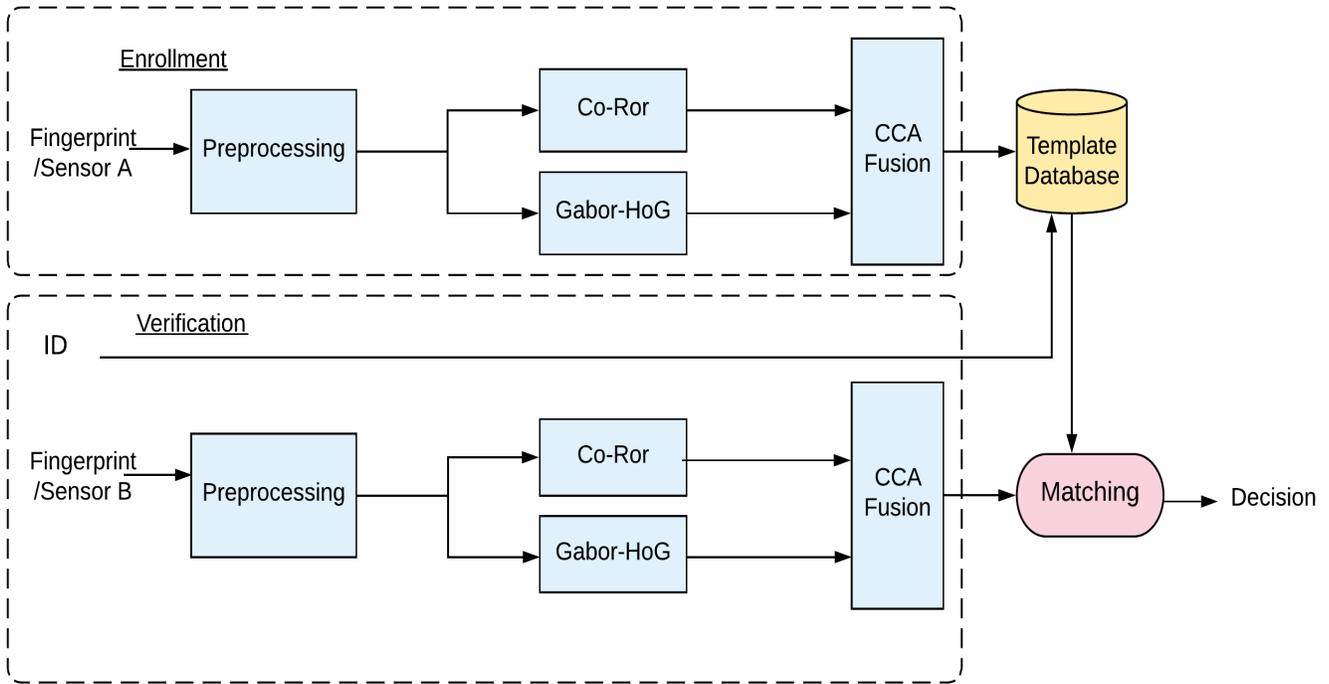

FIGURE 1. An overview of CrossVFinger.

The range of orientation filed is changed from [0, $\pi$] to [0, 180]. The number of different orientations in the orientation field determines the size of the Co-Ror matrix; consequently, the size of a Co-Ror matrix is 180×180=32,400 elements, which not only involves high computationally cost but also is not easily manageable in terms of memory space.

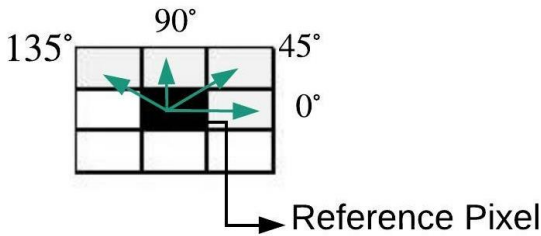

FIGURE 2. Directions of the co-occurrence matrix for extracting the features.

In addition, it embeds noise. To overcome these issues, the orientations are quantized into eight dominant orientations, $\theta_1, \theta_2, \ldots, \theta_n$, as shown in Table 1. This results in a Co-Ror matrix of size, which is easily manageable and suppresses the noise in the orientation filed.

Fig. 3 shows the example computation of a Co-Ror matrix with $d = 1, \phi = 0°, 45°, 90°, 135°$ of a patch of orientation field of a fingerprint. Each element of a Co-Ror is the number of times two orientations $i$ and $j$ coexist along direction ɸ and distance $d$ apart. For example, orientations 1 and 1 co-occur four times along the direction 0 and distance one unit apart, so $h_{1,0°}(1,1) = 4$; similarly, $h_{1,0°}(1,2) = 3$ and so on.

After computing Co-Ror matrices, they are vectorized and concatenated to form a *Co-Ror descriptor*. Finally, the descriptor are normalized to have a zero mean and unit length. The normalization transforms the descriptors into common domain and simplifies subsequent calculations. Fig. 4 shows the process of computing a Co-Ror descriptor with different offsets (ɸ and $d$).

TABLE I
QUANTIZATION OF ORIENTATION VALUES

| Dominant orientations | $\theta_1$ | $\theta_2$ | $\theta_3$ | $\theta_4$ | $\theta_5$ | $\theta_6$ | $\theta_7$ | $\theta_8$ |
|---|---|---|---|---|---|---|---|---|
| Orientations range | $(0°, 22.5°]$ | $(22.5°, 45°]$ | $(45°, 67.5°]$ | $(67.5°, 90°]$ | $(90, 112.5°]$ | $(112.5°, 135°]$ | $(135°, 157.5°]$ | $(157.5°, 180°]$ |
| Quantized value | 1 | 2 | 3 | 4 | 5 | 6 | 7 | 8 |

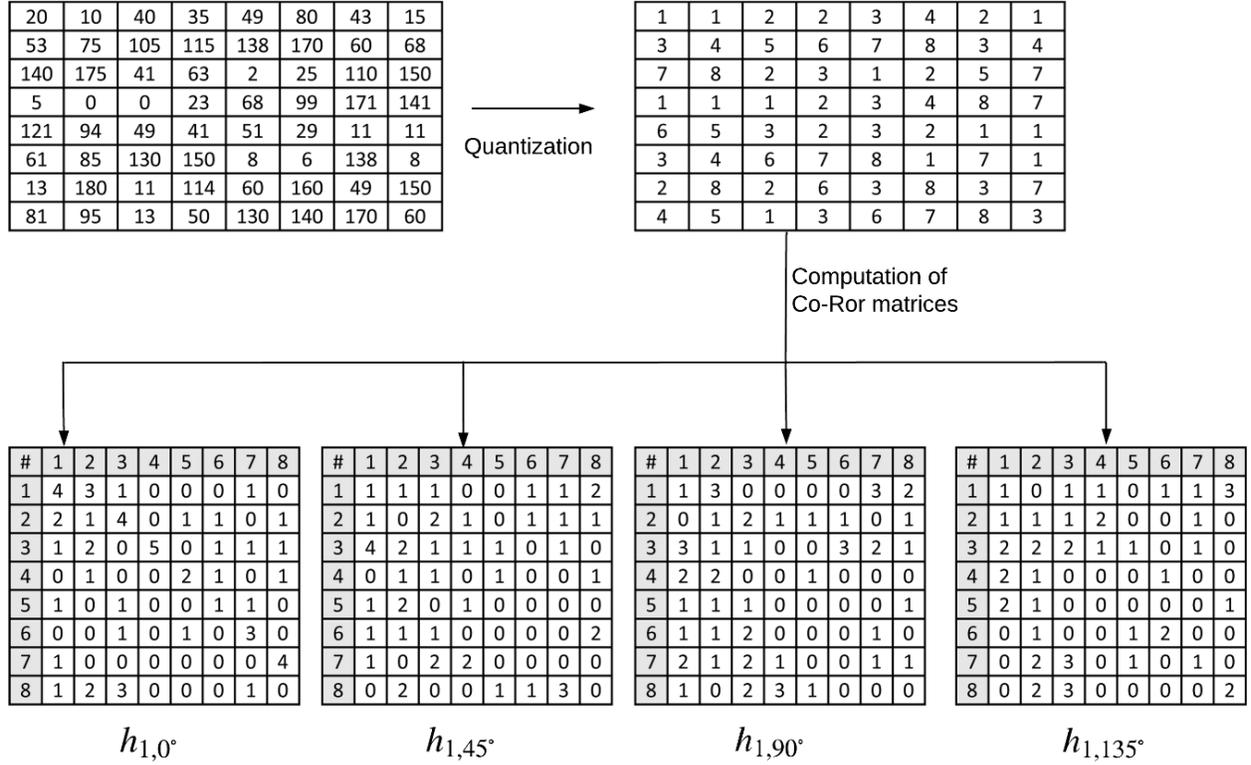

FIGURE 3. The computation of Co-Ror matrices: Top-left: an 8×8 patch a fingerprint orientation field, bottom: four Co-Ror matrices along four directions: $\phi=0°$, $45°$, $90°$, and $135°$.

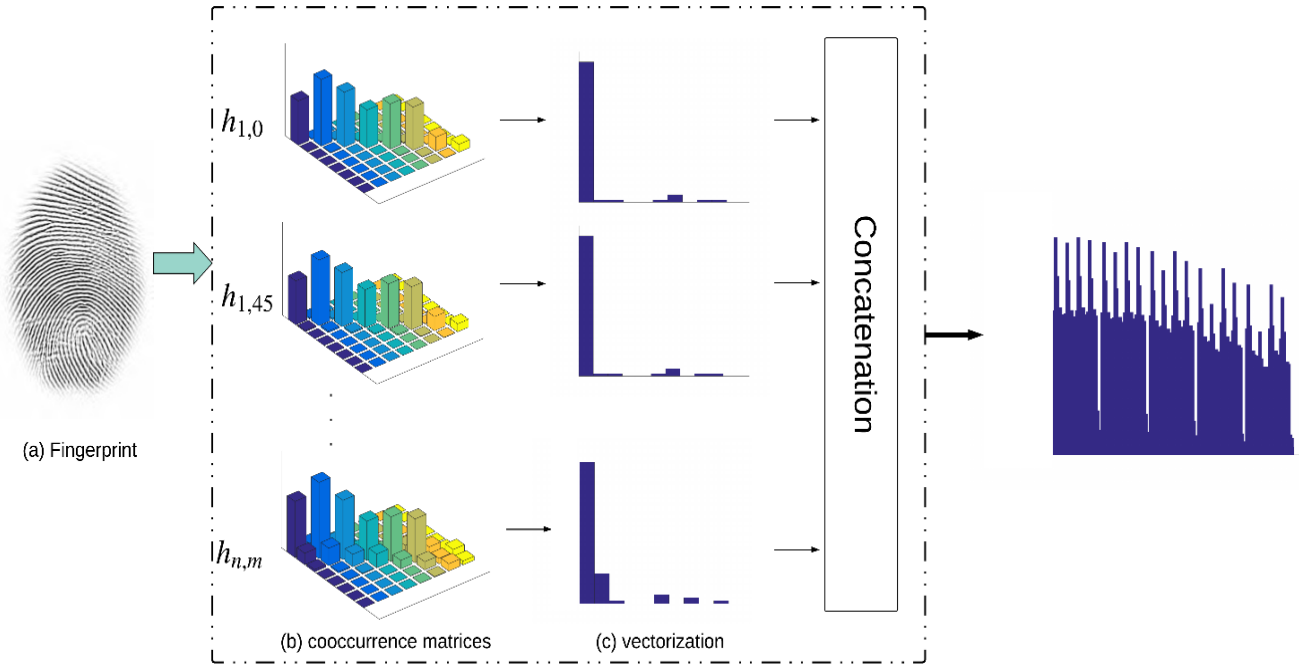

**FIGURE 4. Overview of the computation process of the Co-Ror descriptor.**

A fingerprint contains connected ridges. The distance between ridges is an important visual cue for fingerprint recognition [1, 34], but it offers difficulty when dealing with fingerprint sensor interoperability [12-15]. Figure 5 exhibits four fingerprints and their thinned versions. These fingerprints belong to the same finger and were captured with different sensors; these are taken from the FingerPass database (its detailed account is given in Section IV). The inter-ridge distance is different in the impressions captured with different sensors, see the thinned fingerprints; it causes the rejection of a genuine fingerprint match. Thus, to tolerate the ridge spacing effect, different $d$ distances are adopted to account for the difference in inter-ridge distance.

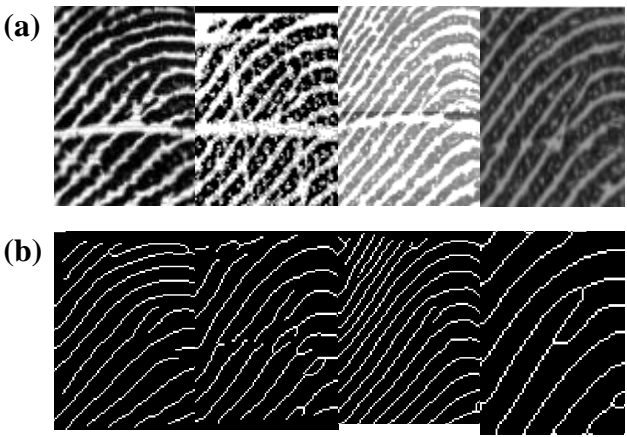

**FIGURE 5. (a) Zoomed-in views of fingerprints of the same finger captured with different sensors and (b) the corresponding thinned versions.**

When impressions of a finger are captured with different sensors, the ridge patterns remain same. These patterns are the most pronounced structural characteristic of a fingerprint and is therefore a strong feature for differentiation. The Co-Ror captures the characteristics of fingerprint ridge patterns, regardless of the details of the local textures and scales. The Co-Ror employs different directions, which lead to capturing the multi-directional relationships of ridge orientation patterns. Moreover, the Co-Ror uses a different values of $d$ depending on the inter-ridge spacing. Thus, we argue that the proposed descriptor has the potential to tackle the fingerprint sensor interoperability problem.

*2)* GABOR-HOG

The Gabor-HoG descriptor is based on the histograms of oriented gradients computed from multi-scale and multi-directional feature maps derived with Gabor filtering; thus it encodes a detailed description of scale based local ridge orientations. The Gabor-HoG descriptor was first employed by Nanni et al. [16] in their fingerprint recognition method and they used four orientations. Unlike Nanni et al., in this work, we used eight orientations. With eight orientations, Gabor-HoG extracts richer information about the scale based local ridge orientations than when four orientations are used.

When constructing the Gabor-HoG descriptor, firstly feature maps are generated by filtering a fingerprint image with the Gabor filter bank comprising four scales and eight orientations ($\theta = 0°, 22.5°, 45°, 67.5°, 90°, 112.5°, 135°, 157.5°$). This process is common practice in the literature of fingerprint recognition [17], [18]. The HoG is then calculated from each feature map using 3×3 cells. The HoG descriptors extracted from all feature maps are normalized to reduce the effect of the variation in gray level values along the furrows and ridges and to suppress artifacts caused due to sensor noise. Finally, the descriptors are concatenated.

### B. FEATURE FUSION USING CANONICAL CORRELATION ANALYSIS (CCA)

The two descriptors (Co_Ror and Gabor-HoG) are extracted from a fingerprint and reflect different fingerprint characteristics and their fusion can result in a robust descriptor. Fusion of the descriptors is expected to improve the system performance, and the fused descriptors hold more information about the fingerprint. The idea is to fuse the descriptors so that the resulting descriptor has maximum correlation with them. This purpose is served by canonical correlation analysis (CCA) [19], which is a statistical method for finding linear relationships between two sets, which have maximum correlation with them. In view of this, we employ CCA for fusion.

CCA [19] has been widely applied to analyze the correlation between two sets of variables. Let $X \in \mathbb{R}^{p \times n}$ and $Y \in \mathbb{R}^{q \times n}$ be two matrices consisting of $n$ training feature vectors corresponding to each of the two descriptors to be fused. Here, $p$ and $q$ refer to the number of features in two descriptors. Let $S_{xx} \in \mathbb{R}^{p \times p}$ and $S_{yy} \in \mathbb{R}^{q \times q}$ denote the within-set covariance matrices of $X$ and $Y$ and $S_{xy} \in \mathbb{R}^{p \times q}$ and $S_{yx} = S_{xy}^T \in \mathbb{R}^{q \times p}$ denote the between-set covariance matrices. The aim of CCA is to compute the projections $X^* = W_X^T X$ and $Y^* = W_Y^T Y$ so that pair-wise correlation between $X^*$ and $Y^*$ is maximum:

$$corr(X^*, Y^*) = \frac{cov(X^*, Y^*)}{var(X^*).var(Y^*)} = \frac{W_x^T S_{xy} W_y}{(W_x^T S_{xx} W_x)(W_y^T S_{yy} W_y)} \quad (4)$$

The optimal solution i.e. the transformation matrices $W_x$ and $W_y$ are computed by solving the eigenvalue equations [19]:

$$S_{xx}^{-1} S_{xy} S_{yy}^{-1} S_{yx} \widehat{W_x} = \Lambda^2 \widehat{W_x}$$
$$S_{yy}^{-1} S_{yx} S_{xx}^{-1} S_{xy} \widehat{W_y} = \Lambda^2 \widehat{W_y}.$$

In each equation, the number of non-zero eigenvalues is $k = rank(S_{xy}) \leq \min(n, p, q)$, and the eigenvalues will be ordered in decreasing order $\lambda_1 \geq \cdots \geq \lambda_k$. The transformation matrices $W_x$ and $W_y$ are composed of the eigenvectors corresponding to largest non-zero eigenvalues.

Based on the idea in [19], feature-level fusion is achieved by either the summation or concatenation of the projections:

$$Z_1 = X^* + Y^* = W_X^T X + W_Y^T Y = \begin{pmatrix} W_x \\ W_y \end{pmatrix}^T \begin{pmatrix} X \\ Y \end{pmatrix} \quad (5)$$

$$Z_2 = \begin{pmatrix} X^* \\ Y^* \end{pmatrix} = \begin{pmatrix} W_X^T X \\ W_Y^T Y \end{pmatrix} = \begin{pmatrix} W_x & 0 \\ 0 & W_y \end{pmatrix} \begin{pmatrix} X \\ Y \end{pmatrix} \quad (6)$$

where $Z_1$ and $Z_2$ are the canonical correlation discriminant features (CCDFs).

We used the concatenation approach defined in (6) in this work, it is justified in Section IV-B(2). The matching score between the gallery and query fingerprints is calculated using city-block distance between the extracted features.

### C. MATCHING ALGORITHMS

The detail of the enrollment module of CrossVFinger is summarized in Algorithm 1 and that of the matching module is summarized in Algorithm 2.

---

**Algorithm 1: Enrollment Module**

**Input:**
    T: Template fingerprint.
    ID: Subject ID.

**Processing:**
    Step-1: Compute C, the Co-Ror descriptor of the template fingerprint.
    Step-2: Compute G, the Gabor-HoG descriptor of the template fingerprint.
    Step-3: Compute FT, the fusion of descriptors C and G using CCA.
    Step-4: Save FT with ID in the template database.

---

**Algorithm 2: Matching Module**

**Input:**
    I: Probe fingerprint.
    ID: Subject ID.

**Output:**
    Score: the matching score.

**Processing:**
    Step-1: Compute C, the Co-Ror descriptor from fingerprint I.
    Step-2: Compute G, the Gabor-HoG descriptor from fingerprint I.
    Step-3: Compute FI, the fusion of the descriptors C and G using CCA.
    Step-4: Retrieve the FT descriptors of r fingerprints of the subject with ID from the template database: T1, T2, .., Tr.
    Step-5: Initialize Si to zero.
    Step-6: **for** i = 1 : r **do**
        Compute similarity score Si = d(FI, FTi).
    **end**
    Step-7: Score = min (S1, S2, . . . , Sr).

---

## IV. EXPERIMENTAL RESULTS AND DISCUSSION

Before the results and discussion, the brief description of datasets used in experiments, evaluation protocol, and model selection are presented.

## A. CROSS-SENSOR FINGERPRINT DATABASES AND EVALUATION PROTOCOL

Experiments were performed on two benchmark cross-sensor databases: Multisensor Optical and Latent Fingerprint (MOLF) [20] and FingerPass [21]. Each database consists of fingerprints, which were acquired with sensors having different technology and interaction types.

The MOLF database consists of three datasets, which were captured with three optical sensors: (1) CrossMatch L-Scan Patrol, (2) Lumidigm Venus IP65 Shell, and (3) Secugen Hamster-IV. The fingerprints were acquired from 100 subjects in two sessions; in each session, 2 independent impressions each of 3 slap prints were captured with CrossMatch, 2 impressions each of 10 fingers were acquired with both Lumidigm, and Secugen. The resolution of each sensor i.e. Lumidigm, Secugen, and CrossMatch is 500 dpi, and the sizes of the captured fingerprints are 352 × 544, 258 × 336 and 1600 × 1500 pixels, respectively. The datasets captured with Lumidigm, Secugen, and CrossMatch sensors are referred to as DB1, DB2 and DB3. Fig. 6 depicts three fingerprints of the same finger acquired with the three sensors. The difference in quality, resolution and noise patterns created by different sensors is obvious in the impressions.

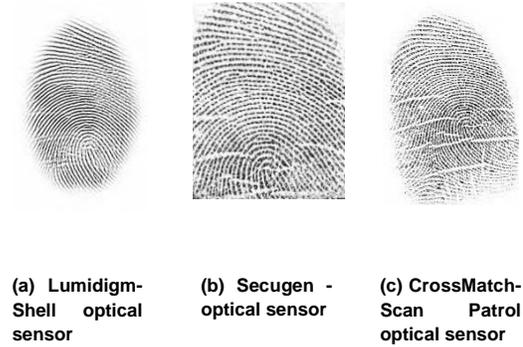

(a) Lumidigm-Shell optical sensor  (b) Secugen - optical sensor  (c) CrossMatch-Scan Patrol optical sensor

**FIGURE 6. Three fingerprints of the same finger from the MOLF database.**

The FingerPass database includes nine datasets acquired with nine sensors of different technology and interaction types. Each dataset consists of 720 fingerprint classes with 12 impressions for each fingerprint class and so the total number of 8,640 fingerprints. The database

TABLE 1.
THE DETAIL OF SENSORS USED TO CAPTURE THE FINGERPASS CROSS-SENSOR DATABASE.

| Sub-dataset | Sensor | Technology Type | Interaction Type | Image Size (pixels) | Image Resolution |
|---|---|---|---|---|---|
| V3O | V300 | Optical | Press | 640×480 | 500 dpi |
| FXO | FX3000 | Optical | Press | 400×560 | 569 dpi |
| URO | URU4000B | Optical | Press | 500×550 | 700 dpi |
| AEO | AES2501 | Optical | Sweep | unfixed | 500 dpi |
| SWC | SW6888 | Capacitive | Sweep | 288×384 | 500 dpi |
| ATC | ATRUA | Capacitive | Sweep | 124×400 | 250 dpi |
| AEC | AES3400 | Capacitive | Press | 144×144 | 500 dpi |
| FPC | FPC1011C | Capacitive | Press | 152×200 | 363 dpi |
| TCC | TCRU2C | Capacitive | Press | 208×288 | 500 dpi |

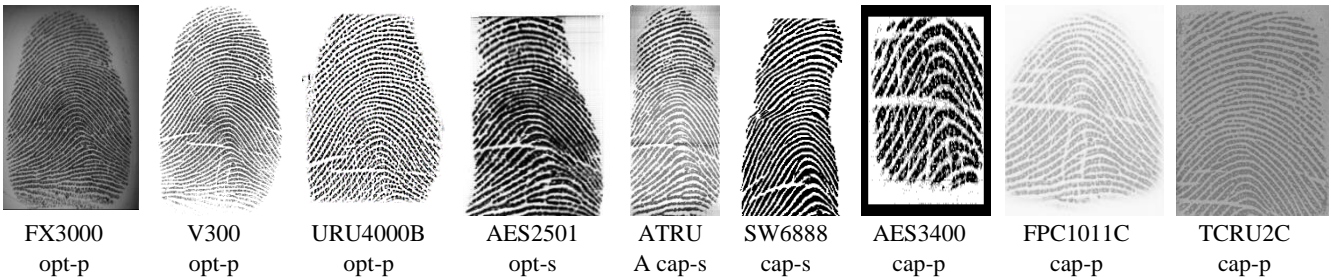

FX3000 opt-p | V300 opt-p | URU4000B opt-p | AES2501 opt-s | ATRUA cap-s | SW6888 cap-s | AES3400 cap-p | FPC1011C cap-p | TCRU2C cap-p

**FIGURE 7. Nine fingerprints of a finger taken from the FingerPass database; opt and cap mean optical and capacitive sensors, respectively; p and s mean press and sweep capture type, respectively.**

consists of 77,760 fingerprints. The detail of sensors is given in Table 2; there are three optical sensors with press, one sensor with sweep, three capacitive sensors with press and two capacitive sensors with sweep interaction type.

Fig. 7 depicts nine fingerprints of a finger selected from the FingerPass database. As is clear from the example impressions, it is a challenging database.

There are two matching scenarios of interest when evaluating a matching system: 1) *regular matching* (also known as intra-device, native-device or simply native matching), two fingerprints acquired with same sensor are compared and the performance metric is termed as native equal error rate (native-EER); and 2) *cross-matching* (also called inter-device, cross-device or cross-sensor matching), in this case two fingerprints acquired with different sensors are compared for verification and the performance metric is known as interoperable or cross-EER. For matching, we used the same evaluation protocol as was adopted by Jia et al. [21] to divide the data into gallery and query sets, and to compute genuine match scores and impostor match scores. For cross-matching scenario, the gallery set consists of the fingerprints captured with one sensor and the query set contains the fingerprints acquired with the other sensor.

Two fundamental metrics for evaluating a matching method are the false match rate (FMR) and false non-match rate (FNMR). For evaluating the performance of the proposed method and comparing it with the stat-of-the art methods, we employed the well-known metric i.e. equal error rate (EER), which is commonly used in authentication scenarios; it is the operating point where two fundamental metrics i.e. the false match rate (FMR) and false non-match rate (FNMR) are equal. In addition, we used another comprehensive metric i.e. the detection error tradeoff (DET) curve, which plots FMR vs. FNMR.

### B. MODEL SELECTION

The CrossVFinger includes various parameters, which effect its performance and their suitable choice is essentail for its best performance. In the sequel, the effects of the parameters on authentication performance have been discussed, and the best choices for them have been suggested.

The CrossVFinger was implemented in the Matlab (R2016a) environment, and the experiments were performed on a PC (Intel Core i7-4702MQ processor, 2.2 GHz, 4 cores) with 14 GB RAM and the Microsoft Windows 10 x64 operating system.

#### 1) AN ANALYSIS OF THE EFFECT OF THE CO-ROR PARAMETERS

The Co-Ror involves two parameters, $\phi$ and $d$; three configurations of $\phi$ were tested to select the best configuration for $\phi$: $\phi = 2$ ($\theta = 0°, 90°$), $\phi = 4$ ($\theta = 0°, 45°, 90°, 135°$) and $\phi = 8$ ($\theta = 0°, 22.5°, 45°, 67.5°, 90°, 112.5°, 135°, 157.5°$).
Additionally, four metrics were employed for computing the similarity score for matching: Euclidean distance, city-block distance, histogram intersection, and chi-square distance.

Typically, $d$ takes an integer value and can be in any range selected from the set of the integers. Fig. 8 shows box plots of the inter-ridge distances for each dataset of the FingerPass database. The ridge spacing is in the range [5-11]. We argue that choosing the value of $d$ to reflect the inter-ridge distance will improve the robustness. To assess the effects of $d$, we examined two configurations with fixed integer values: c1=(1,2,3) and c2=(1,2,3,4) and two configurations where the distance depends on the inter-ridge spacing: g1= (d, 2d, 3d) and g2 =(d, 2d, 3d, 4d). We chose $d$ to be the inter-ridge spacing of the fingerprint image. Instead of using the precise inter-ridge distance, which is time consuming, we fix it to 5 based on the observation from Fig. 8. To tackle the problem of variation in inter-ridge distance of cross-sensor fingerprints, we compute Co-Ror descriptor choosing more than one values of d.

Fig. 9 depicts the results of Co-Ror descriptor on two datasets (B1 and B2) of the MOLF database. The dataset B1 acquired with the Lumidigm sensor was employed for enrollment whereas the dataset B2 acquired with the Secugen sensor was employed for authentication. The results indicate that the configurations that depend on inter-ridge distance are better than those using fixed distances.Within the configurations that depend on inter-ridge distance, there is no significant difference in EER values. Therefore, we chose g1 = (d, 2d, 3d), where d = 5, to reduce the computation required by the proposed descriptor. Furthermore, Fig. 9 shows that Φ=4 generates results, which are better than those of other configurations. Moreover, the results show that city-block distance is the best matching metric. The performances of Euclidean distance and chi-square distance are worse than those of the city-block distance, and the histogram intersection distance yields the poorest performance.

The above results and discussion indicate that the best choice for the parameters is ($\theta = 0°, 45°, 90°, 135°$) for direction $\phi$, g1=(d,2d,3d) with d=5 for offset, and city-block distance for matching.

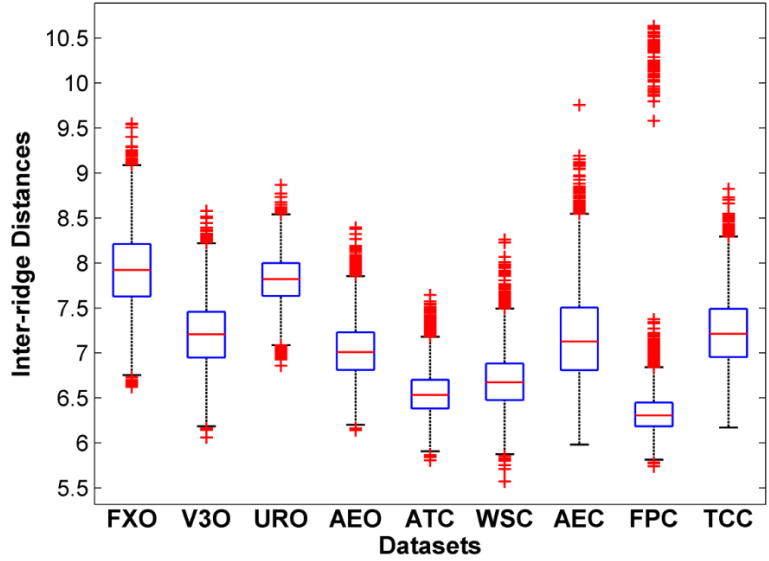

**FIGURE. 8. Box plot of inter-ridge distances.**

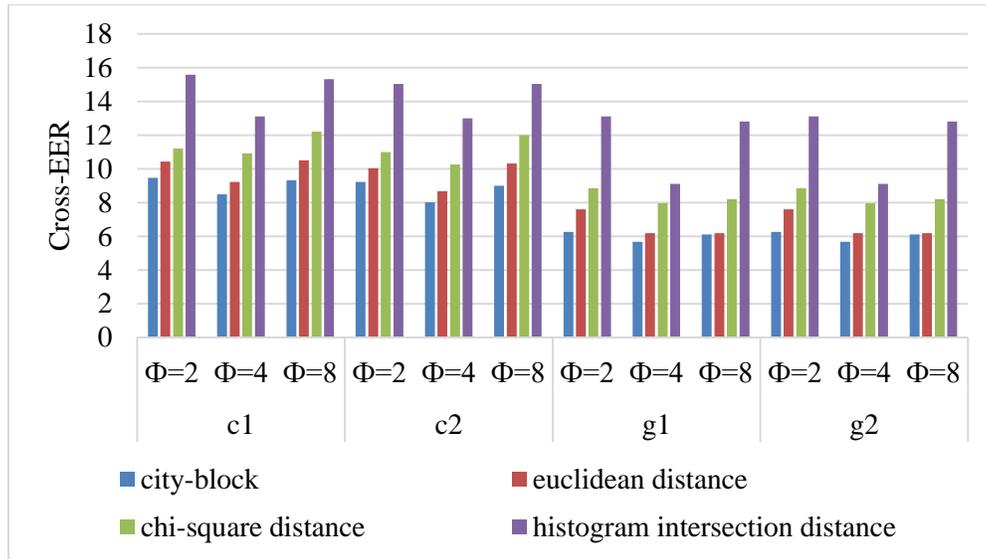

**FIGURE 9. Mean cross-EER showing the effects of parameters (the number of directions and offset) of Co-Ror descriptor and similarity measures.**

2) EFFECT OF CCA FUSION TYPE

Feature-level fusion using CCA is achieved by either the summation or concatenation of the projected feature vectors, see the (5) and (6). To select the best type of fusion, we performed an experiment using the same datasets and the parameters selected in the previous section. Fig. 10 shows that feature fusion via concatenation produces better results in terms of EER than those of summation fusion. This observation suggests the adoption of concatenation for CCA fusion.

## C. RESULTS AND DISCUSSION

We performed extensive experiments on two databases. This section presents the performance of CrossVFinger on the two databases.

1) EXPERIMENTAL RESULTS ON THE MOLF DATABASE

We performed three sets of cross matching experiments on the MOLF database. Two sets of experiments were performed with the Co-Ror and Gabor-HoG descriptors without fusion to show the effectiveness of each individual descriptor. The third set of experiments was performed by fusing the two descriptors with CCA fusion method (i.e., the proposed CrossVFinger method). Table 3 shows the cross matching results in terms of EER. Though each descriptor performs relatively better for native matching than cross matching, the overall performance of Gabor-HoG is worse than that of Co-Ror. Moreover, CCA fusion results in a significant performance improvement. The reason is that CCA extracts the discriminative information from both descriptors and suppresses the redundancy in each descriptor by maximizing the correlation among the fused descriptors.

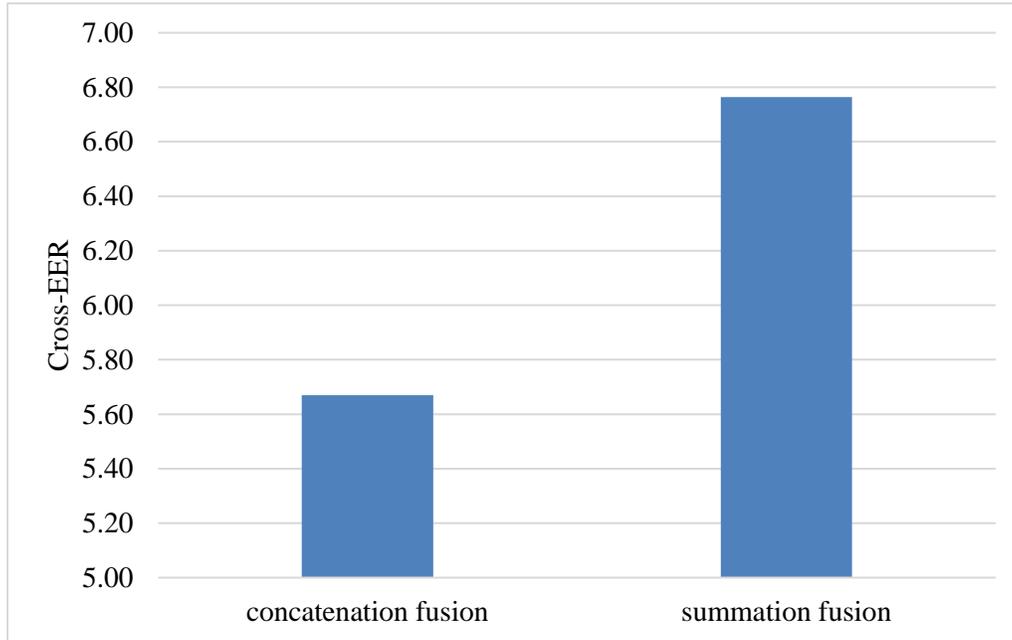

**FIGURE 10. Average cross-EER of the CCA fusion showing the effect of fusion type.**

TABLE 2.
THE RESULTS IN TERMS OF EER OF (A) CO-ROR, (B) GABOR-HOG, AND (C) CROSSVFINGER ON THE MOLF DATABASE.

| Gallery→ Probe | (a) Co-Ror descriptor | | | (b) Gabor-HoG descriptor | | | (c) CrossVFinger | | |
|---|---|---|---|---|---|---|---|---|---|
| | DB1 | DB2 | DB3 | DB1 | DB2 | DB3 | DB1 | DB2 | DB3 |
| DB1 | **2.90** | 3.64 | 4.49 | **6.80** | 11.81 | 9.93 | **0.31** | 1.85 | 1.03 |
| DB2 | 3.64 | **1.53** | 2.90 | 11.81 | **7.48** | 8.79 | 1.85 | **0.51** | 1.63 |
| DB3 | 4.49 | 2.90 | **2.25** | 9.93 | 8.79 | **6.59** | 1.03 | 1.63 | **0.48** |

### 2) EXPERIMENTAL RESULTS ON THE FINGERPASS DATABASE

Table 4 shows the verification results of CrossVFinger in terms of EER on the FingerPass database. The native EER is relatively small and is less than 1 for all sensors. Though, for cross-matching cases, the cross-EERs are slightly higher than the native EERs, they are very small in most of the cases. The cross-EER is highest among all cross-sensor cases when AEC and FPC sensors are used for the probe or gallery. AEC and FPC are both capacitive, press interaction sensors, with image sizes of 144×144 pixels and 152×200, respectively. The reason for the poor performance when AEC and FPC are used for the gallery or probe is likely due to very low resolution of the acquired fingerprint images. When probe and gallery sensors are of optical type (FXO, V3O, URO, and AEO), in cross-matching cases, the cross-EER is small (all less than 1).

When ATC, AEC, or FPC (capacitive-type sensors) are used for the probe or gallery, the cross-EER is higher because the fingerprint images have low resolution than those by optical sensors.

If ATC, AEC, or FPC are used for the galley or probe, the cross-EER is greater than 1 in most cases. A closer look at the image resolutions and sizes of the corresponding fingerprints obtained from ATC, FPC, and AEC reveals a possible correlation between cross-EER and image resolution and size; the lower the fingerprint resolution or size, the higher the cross EER. For best performance, the resolution must be at least 500dpi and the size must be 500x500 pixels.

Fig. 11 shows the average cross-EERs of CrossVFinger on the datasets of the FingerPass database. The interoperable EER for all datasets in FingerPass is less than 3.5%, except AEC and FPC datasets; average cross EERs for AEC and FPC are significantly higher for the reasons discussed above. Overall, the performance shows that the adopted descriptor is robust in encoding the distribution of ridge patterns, and the CrossVFinger provides outstanding results when the image resolution is not too low and the image size is not too small. Conversely, the performance decreases when fingerprints have low resolutions or small sizes, as in the case of the AEC and FPC datasets.

TABLE 3.
VERIFICATION RESULTS OF THE PROPOSED METHOD IN TERMS OF EER ON THE FINGERPASS DATABASE.

| (Template/Probe) | FXO | V3O | URO | AEO | ATC | SWC | AEC | FPC | TCC |
|---|---|---|---|---|---|---|---|---|---|
| FXO | **0.008** | 0.028 | 0.783 | 0.357 | 1.378 | 0.661 | 4.363 | 1.397 | 0.271 |
| V3O | 0.028 | **0.013** | 0.758 | 0.392 | 1.165 | 0.270 | 4.975 | 6.648 | 0.243 |
| URO | 0.783 | 0.758 | **0.006** | 0.677 | 1.016 | 0.771 | 5.565 | 6.829 | 0.247 |
| AEO | 0.357 | 0.392 | 0.677 | **0.005** | 1.268 | 0.777 | 6.543 | 6.872 | 0.684 |
| ATC | 1.378 | 1.165 | 1.016 | 1.268 | **0.305** | 0.452 | 6.717 | 1.580 | 0.841 |
| SWC | 0.661 | 0.270 | 0.771 | 0.777 | 0.452 | **0.002** | 6.427 | 1.593 | 0.446 |
| AEC | 4.363 | 4.975 | 5.565 | 6.543 | 6.717 | 6.427 | **0.578** | 6.471 | 1.086 |
| FPC | 1.397 | 6.648 | 6.829 | 6.872 | 1.580 | 1.593 | 6.471 | **0.754** | 5.877 |
| TCC | 0.271 | 0.243 | 0.247 | 0.684 | 0.841 | 0.446 | 1.086 | 5.877 | **0.039** |

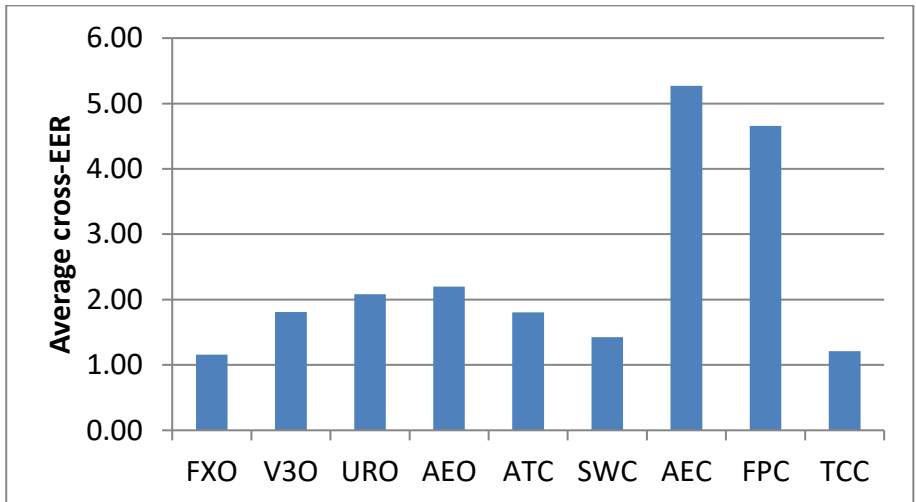

**FIGURE 11. Average cross-EERs of CrossVFinger on the datasets of the FingerPass database.**

The FingerPass database has the capacity to be used to analyze the effectiveness of CrosseVFinger on the technology type of sensors. Based on sensor technology type, the FingerPass can be grouped into two categories: capacitive and optical. The optical group includes datasets captured with four sensors from FXO to AEO, and the capacitive group contains datasets acquired with the rest of five sensors. Fig. 12 shows the average interoperable EER of CrossVFinger according to sensor technology type. The sensor-EER (when different sensors but of the same technology type are used for enrolment and verification, i.e., capacitive vs. capacitive and optical vs. optical) of each type is calculated as the average of the cross-EERs of the same sensor type from Table 4, whereas the cross-sensor EER (when sensors of different technology types are used for enrolment and verification, i.e., optical vs. capacitive and vice versa) is computed as the average of the cross-EERs obtained when sensors of divergent types are used. According to Fig. 12, higher sensor-EER is achieved by the capacitive vs. capacitive group compared to that of the optical vs. optical group; for optical type it is less than 0.5%. The poor average performance of capacitive sensors is due to AEC and FPC, which generate impressions of either low resolution or small sizes.

### D. COMPARISONS WITH THE STATE-OF-THE-ART METHODS

To validate the efficacy of CrossVFinger, its performance is compared with four fingerprint matching methods: VeriFinger [23], MCC [22], MCC+Scale [7], TPS [4], and CrossSFmatching [24]. VeriFinger is a commercial fingerprint matching method developed by Neurotechnology. MCC is a minutia-based matching algorithm, whereas MCC+Scale is an enhanced version of MCC. The CrossSFmatching is based on encoding the fingerprint discriminative features using two minutiae based descriptors and an orientation descriptor. MCC and VeriFinger are considered by various researchers to be the baseline for comparisons for cross-matching and regular matching [7], [21]. In this study, we employed VeriFinger Extended SDK 9.0 and MCC SDK Version 2.0.

1) RESULTS ON MOLF DATABASE

Table 5 reports the performance of CrosseVFinger, VeriFinger , MCC , and CrossSFmatching on the MOLF database. In general, the cross-EER is higher than the native EER for all methods. Overall, the MCC yields poor results, whether regular matching or cross matching; however, the performance is the worst for cross matching scenario. Although VeriFinger produces better results than those of MCC, it also yields poor results for cross matching scenario. CrossSFmatching and Co-Ror descriptor outperform MCC and VeriFinger. This demonstrates the potential of the proposed descriptor to extract discriminative features for a fingerprint matching algorithm.

The CrossVFinger achieves lower cross-EER and native EER compared to MCC and VeriFinger for all three datasets of the MOLF database. Moreover, CrossVFinger outperforms CrossSFmatching except for DB2 vs. DB3 and vice versa. Figures 13, 14, and 15 show the DET curves for the four methods and the proposed Co-Ror descriptor. The DET curves are almost in agreement with the results in Table 5. The CrossVFinger always stands out in terms of DET curves, and the difference is notable. Moreover, the Co-Ror descriptor alone outperforms VeriFinger and MCC.

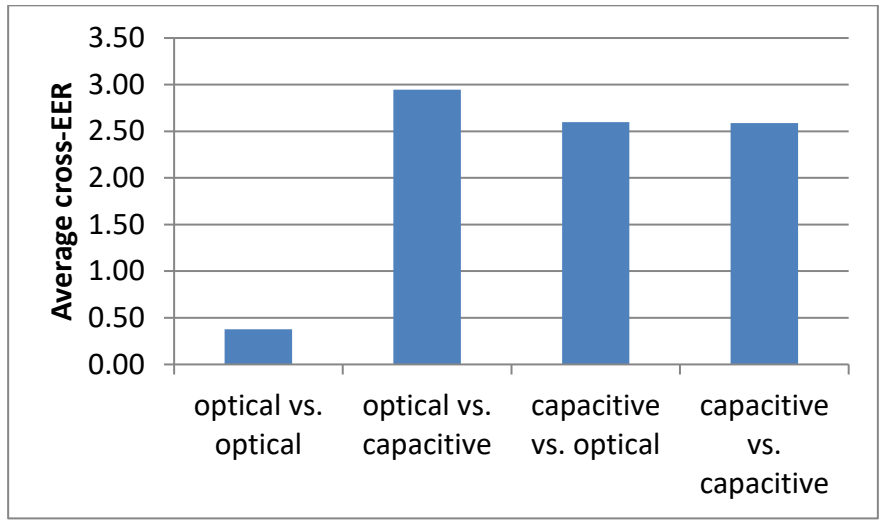

**FIGURE 12.** Average sensor EERs (opt. vs. opt. and cap. vs. cap.) and cross-sensor EERs (opt. vs. cap and vice versa) of CrossVFinger.

TABLE 4.
VERIFICATION RESULTS IN TERMS OF EER OF FIVE METHODS ON THE MOLF DATABASE.

| Gallery→ | (a) MCC Method | | | (b) VeriFinger Method | | | (c) CrossSFmatching | | | (d) Co-Ror Descriptor | | | (e) CrossVFinger | | |
|---|---|---|---|---|---|---|---|---|---|---|---|---|---|---|---|
| Probe | DB1 | DB2 | DB3 | DB1 | DB2 | DB3 | DB1 | DB2 | DB3 | DB1 | DB2 | DB3 | DB1 | DB2 | DB3 |
| DB1 | 11.14 | 18.48 | 20.81 | 3.16 | 6.46 | 6.42 | 0.60 | 1.99 | 1.19 | 2.90 | 3.64 | 4.49 | 0.31 | 1.85 | 1.03 |
| DB2 | 18.48 | 16.82 | 22.74 | 6.47 | 3.2 | 3.94 | 1.99 | 0.64 | 1.24 | 3.64 | 1.53 | 2.90 | 1.85 | 0.51 | 1.63 |
| DB3 | 20.81 | 22.74 | 13.83 | 6.42 | 3.94 | 3.51 | 1.19 | 1.24 | 0.54 | 4.49 | 2.90 | 2.25 | 1.03 | 1.63 | 0.48 |

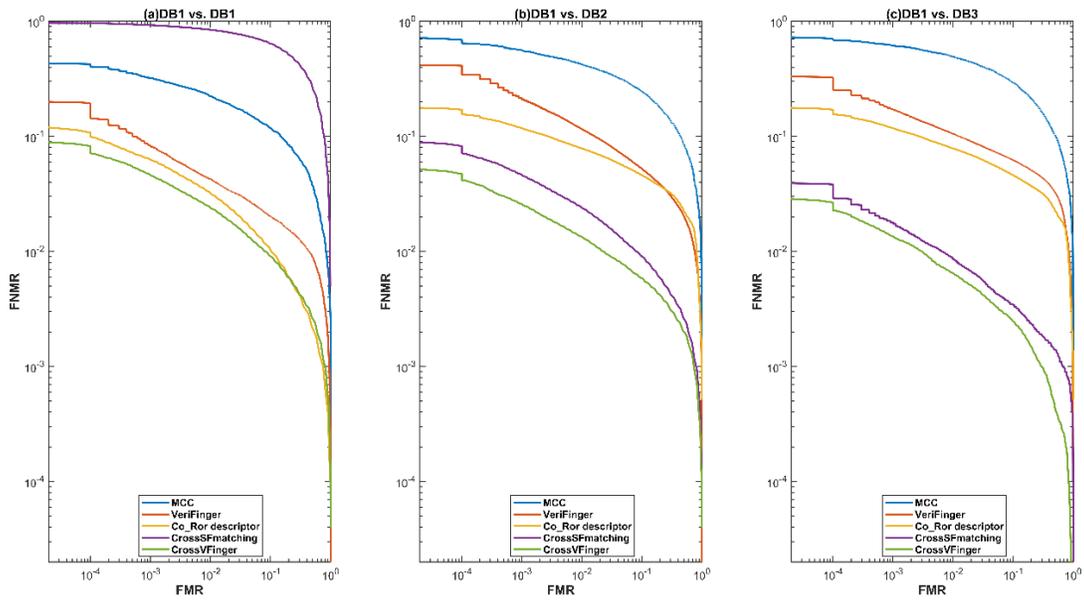

**FIGURE 13. DET curves corresponding to the four methods and Co_Ror based method the MOLF database, DB1 is used as gallery.**

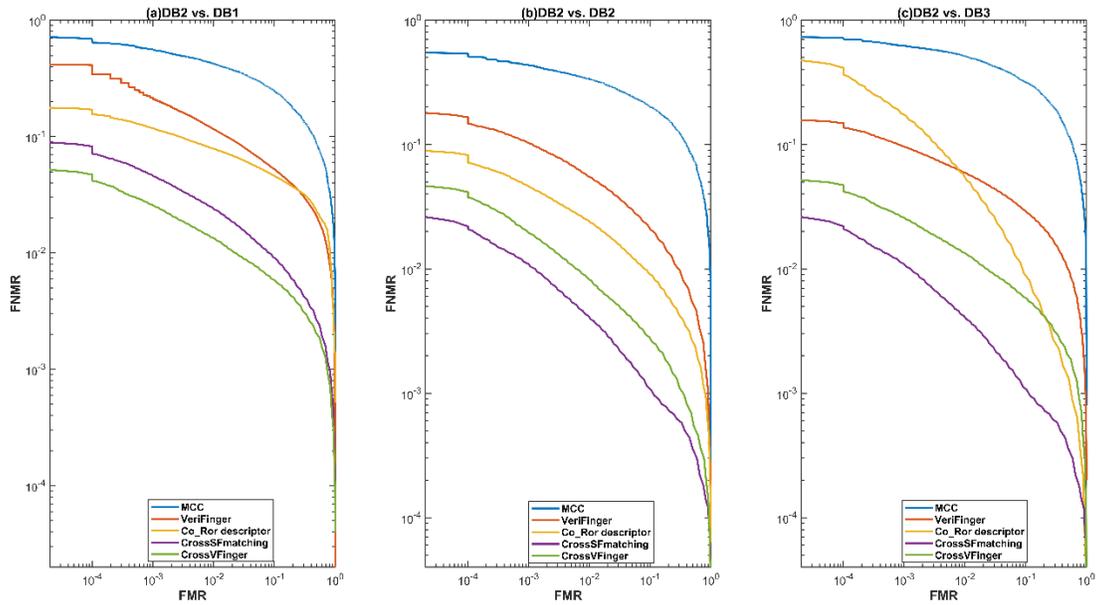

**FIGURE 14. DET curves corresponding to the four methods and Co_Ror based method on the MOLF database, DB2 is used as gallery.**

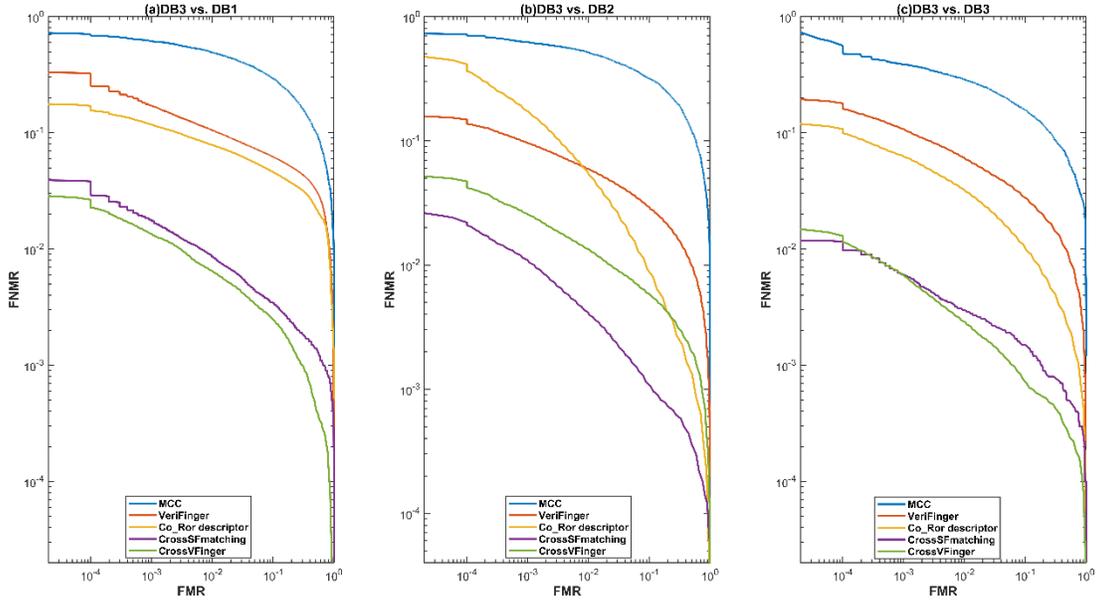

**FIGURE 15.** DET curves corresponding to the four methods and Co_Ror descriptor on the MOLF database, DB3 is used as gallery.

Table 6 shows the average matching times taken by different methods on the MOLF database. CrossVFinger is faster than VeriFinger, MCC, and CrossSFmatching. The reason that CrossVFinger takes less time is that it is based on an alignment-free approach.

### 2) RESULTS ON THE FINGERPASS DATABASE

The cross matching results reported in [7] have been obtained on four subsets selected from the FingerPass: URO (optical, press), TCC (capacitive, press), AEO (optical, sweep), and SWC (capacitive, sweep), i.e. two optical and two capacitive with press and sweep interaction types, all resulting in fingerprints of resolutions 500 dpi or above. We evaluated CrossVFinger on these datasets to compare it fairly with the state-of-the-art methods. Table 7 presents the results in terms of EER, FMR100, FMR1000, and ZeroFMR on the four datasets.

TABLE 5.
AVERAGE TIMES (IN MILLISECONDS) ON THE MOLF DATABASE.

| Methods | VeriFinger | MCC | CrossSFmatching | CrossVFinger |
|---|---|---|---|---|
| Matching time | 0.794 | 0.0238 | 1.943 | **0.000143** |

TABLE 6.
THE RESULTS OF CROSSVFINGER AND FIVE STATE-OF-THE-ART METHODS ON FOUR DATASETS FROM THE FINGERPASS DATABASE.

| Template | Probe | Method | EER (%) | FMR100 (%) | FMR1000 (%) | ZeroFMR (%) |
|---|---|---|---|---|---|---|
| **URO** | URO | MCC | 0.023 | 0.007 | 0.019 | 0.858 |
| | | VeriFinger | 0.018 | - | - | 0.421 |
| | | CrossSFmatching | 0 | 0 | 0 | 0 |
| | | **Co-Ror** | 0.011 | 0.013 | 0.021 | 0.020 |
| | | **CrossVFinger** | 0.006 | 0.008 | 0.008 | 0.013 |
| **TCC** | TCC | MCC | 0.056 | 0.015 | 0.047 | 1.229 |
| | | VeriFinger | 0.045 | - | - | 0.137 |
| | | CrossSFmatching | **0** | **0** | **0** | **0** |
| | | **Co-Ror** | 0.056 | 0.048 | 0.0694 | 0.076 |
| | | **CrossVFinger** | **0.039** | **0.042** | **0.053** | **0.078** |
| **AEO** | AEO | MCC | 0.053 | 0.017 | 0.044 | 1.191 |
| | | VeriFinger | 0.014 | - | - | 0.042 |
| | | CrossSFmatching | 0.05 | 0 | 0.0909 | 2.45 |
| | | **Co-Ror** | 0.009 | 0.008 | 0.015 | 0.021 |
|  |  | **CrossVFinger** | **0.005** | **0.004** | **0.008** | **0.0105** |
| **SWC** | SWC | MCC | 0.073 | 0.028 | 0.061 | 2.517 |
|  |  | VeriFinger | 0.028 | - | - | 0.109 |
|  |  | CrossSFmatching | 0.05 | 0.1818 | 3.909 | 1.909 |
|  |  | **Co-Ror** | 0.048 | 0.048 | 0.061 | 0.067 |
|  |  | **CrossVFinger** | **0.002** | **0.003** | **0.004** | **0.004** |
| **URO** | TCC | MCC | 27.41 | 89.23 | 98.65 | 99.99 |
|  |  | MCC+Scale | 0.283 | 0.128 | 0.457 | 2.936 |
|  |  | TPS | 0.298 | 0.128 | 0.512 | 1.552 |
|  |  | VeriFinger | 0.272 | - | - | 0.714 |
|  |  | CrossSFmatching | 0.82 | 0.517 | 1.63 | 3.014 |
|  |  | **Co-Ror** | 1.834 | 2.085 | 2.669 | 3.434 |
|  |  | **CrossVFinger** | **0.246** | **0.160** | **0.351** | **0.358** |
| **AEO** | SWC | MCC | 2.482 | 3.67 | 7.681 | 24.93 |
|  |  | MCC+Scale | 2.242 | 3.309 | 5.827 | 11.42 |
|  |  | TPS | 2.017 | 3.546 | 7.328 | 15.63 |
|  |  | VeriFinger | 1.083 | - | - | 3.511 |
|  |  | CrossSFmatching | 0.67 | 0.003 | 0.017 | 0.545 |
|  |  | **Co-Ror** | 1.022 | 1.023 | 1.36 | 1.79 |
|  |  | **CrossVFinger** | **0.773** | **0.773** | **1.010** | **1.077** |
| **URO** | AEO | MCC | 27.97 | 90.01 | 98.44 | 99.97 |
|  |  | MCC+Scale | 2.432 | 4.186 | 10.89 | 31.06 |
|  |  | TPS | 2.288 | 3.747 | 10.42 | 27.69 |
|  |  | VeriFinger | 2.675 | - | - | 6.631 |
|  |  | CrossSFmatching | 0.55 | 1.090 | 1.182 | 4.818 |
|  |  | **Co-Ror** | 1.407 | 1.494 | 1.925 | 2.685 |
|  |  | **CrossVFinger** | **0.676** | **0.509** | **2.321** | **2.331** |
| **TCC** | AEO | MCC | 3.305 | 5.358 | 10.5 | 33.1 |
|  |  | MCC+Scale | 2.632 | 3.581 | 6.137 | 15.26 |
|  |  | TPS | 1.948 | 2.444 | 5.758 | 8.683 |
|  |  | VeriFinger | 2.907 | - | - | 8.159 |
|  |  | CrossSFmatching | 0.18 | 0 | 1.091 | 1.727 |
|  |  | **Co-Ror** | 1.255 | 1.319 | 1.926 | 2.170 |
|  |  | **CrossVFinger** | **0.683** | **0.454** | **0.831** | **0.966** |
| **URO** | SWC | MCC | 26.41 | 87.26 | 97.39 | 99.86 |
|  |  | MCC+Scale | 3.326 | 7.854 | 15.47 | 28.37 |
|  |  | TPS | 3.158 | 7.329 | 13.56 | 25.73 |
|  |  | VeriFinger | 3.487 | - | - | 10.92 |
|  |  | CrossSFmatching | 2.74 | 0 | 1.636 | 6.01 |
|  |  | **Co-Ror** | 3.317 | 4.330 | 5.305 | 6.675 |
|  |  | **CrossVFinger** | **0.771** | **0.671** | **5.684** | **5.791** |
| **TCC** | SWC | MCC | 5.21 | 9.701 | 18.59 | 47.19 |
|  |  | MCC+Scale | 4.437 | 8.43 | 13.86 | 25.41 |
|  |  | TPS | 4.382 | 8.592 | 13.94 | 25.83 |
|  |  | VeriFinger | 4.263 | - | - | 19.58 |
|  |  | CrossSFmatching | 0.41 | 1.1667 | 4.833 | 17.83 |
|  |  | **Co-Ror** | 2.235 | 2.537 | 3.192 | 3.670 |
|  |  | **CrossVFinger** | **0.328** | **0.259** | **0.349** | **0.446** |

The impact of adopting different sensors for the probe and gallery on the performance of the compared methods is obvious. The proposed descriptor (Co-Ror) outperforms MCC , VeriFinger, and TPS in all except three cases (TCC vs. TCC, SWC vs. SWC, and URO vs. TCC). The proposed method (CrossVFinger) outperforms MCC , VeriFinger, and TPS in all cases of native matching as well as in the cross-sensor matching scenarios by a large margin. CrossVFinger outperforms CrossSFmatching in all cases of native and cross matching except for (URO vs. URO, TCC vs. TCC, AEO vs. SWC, URU vs. AEO, and TCC vs. AEO).

The VeriFinger is based on minutia points, which are computed with proprietary algorithms like ridge count. The TPS method employs thin-spline model to register a pair of fingerprints. MCC encodes the neighborhood of fixed size around a minutia with a cylinder whose height and base reflect the directional and spatial information, respectively. The modified MCC with scale incorporates scaling on fingerprints in MCC method. The empirical results reported in Table 7 reveal that these methods are not effective for cross matching problem; these methods do not explicitly exploit the fingerprint characteristics, which are invariant to sensor technology types. Across different sensors, the fingerprints of a finger include same ridge orientation patterns, which vary in detail like local micro-structures, rotation and scale. The structural patterns, which are not effected by sensor type, must be taken into account when designing a cross matching method, but the designs of the methods discussed above do not draw on this kind of information. On the other hand, CrossVFinger takes into consideration this information through the usage of Co-Ror and Gabor-HoG, and gives better performance.

The CrossVFinger significantly overcomes the effects of fingerprint sensor interoperability; overall, it yields better cross matching performance than VeriFinger, MCC, TPS, MCC with scale, and CrossSFmatching. The results corroborate the potential of CrossVFinger in dealing with the fingerprint sensor interoperability problem because it draws on the descriptors, which are robust to variability in fingerprints caused by the use of different sensors.

Though CrossVFinger excels other methods. Overall the best authentication performance of CrossVFinger was achieved in the scenarios of optical vs. optical, optical vs. capacitive and capacitive vs. capacitive, which result in fingerprints of high resolution. Therefore, we recommend using sensors of optical or capacitive technological type generating impressions of resolution at least 500 dpi for the best verification results.

## V. CONCLUSION

We introduced an automatic fingerprint verification method – CrossVFinger – for cross matching problem. The method is based on a proposed new fingerprint descriptor called the Co-Ror that encodes the spatial relationship of fingerprint ridge orientations. In addition, CrossVFinger draws on the Gabor-HoG descriptor to encode multiscale ridge orientations. These descriptors are fused using CCA, and the similarity scores are computed using city-block distance. The CrossVFinger does not require registration of minutia points, which is an essential step in many state-of-the-art methods. Additionally, CrossVFinger is capable to tackle sensor-dependent structural variability. Its performance has been validated on two benchmark public domain databases, namely, MOLF and FingerPass, which were developed for designing algorithms for fingerprint sensor interoperability problem; the comparison has been made with five state-of-the-art methods: VeriFinger, MCC, TPS, MCC with scale and CrossSFmatching; CrossVFinger significantly outperforms these methods. This study recommends using sensors of optical or capacitive technological type generating impressions of resolution at least 500 dpi for the best verification results in cross matching scenario.

The design of the proposed Co-Ror descriptor is based on the observation that the distribution of ridge orientation patterns doesn't vary significantly in fingerprints captured with different sensors, whereas the inter-ridge distance in fingerprints varies with the sensor type. Extensive experiments on benchmark databases validate the effectiveness of the proposed descriptor. The concrete visualization or reasoning on how it overcomes the cross-sensor characteristics such as sensor type-dependent deformation, scale variation, and partial acquisitions of fingerprints is a future work. In addition, this work is based on the hypothesis that one type of sensor is employed for enrolment and sensor of another kind is installed for authentication. The future work is to investigate the cross matching problem in the scenario when more than one sensors of varying technology and press types are employed for enrolment and a sensor of different kind is used for authentication.

Security agencies, service providers and law-enforcement departments will benefit from this research. Usually the fingerprint databases are enrolled with fingerprint sensors of a specific technology and interaction types; the same type of sensors are used for query. With the advances in fingerprint technology, when the old sensors are replaced with new types for query, it is not only financially demanding but also not easy to manage to replace the enrolled databases with new sensor types. The proposed research will help security agencies, service providers and law-enforcement departments to overcome this problem.